\PassOptionsToPackage{table}{xcolor}
\documentclass[sigconf]{acmart}
\usepackage{latexsym}
\usepackage{subfigure}
\usepackage{amssymb}
\usepackage{amsmath}
\usepackage{stackrel}
\usepackage{booktabs}
\usepackage{makecell}
\usepackage{bm}
\usepackage{color}
\usepackage[table]{xcolor}
\usepackage{multirow}
\usepackage{tcolorbox}
\usepackage{listings}
\usepackage{array}
\usepackage{rotating}
\AtBeginDocument{%
  }

\setcopyright{acmlicensed}
\copyrightyear{2025}
\acmYear{2025}
\setcopyright{cc}
\setcctype{by}
\acmConference[WWW Companion '25]{Companion Proceedings of the ACM Web
Conference 2025}{April 28-May 2, 2025}{Sydney, NSW, Australia}
\acmBooktitle{Companion Proceedings of the ACM Web Conference 2025 (WWW
Companion '25), April 28-May 2, 2025, Sydney, NSW, Australia}
\acmDOI{10.1145/3701716.3715457}
\acmISBN{979-8-4007-1331-6/25/04}

\begin{document}

\title{MAQInstruct: Instruction-based Unified Event Relation Extraction}

\author{Jun Xu}
\orcid{0000-0001-9565-6106}
\affiliation{%
  \institution{AntGroup}
  \city{Hangzhou}
  \country{China}}
\email{xujun.xj@antgroup.com}

\author{Mengshu Sun}
\orcid{0000-0003-2639-9462}
\authornote{Corresponding author.}
\affiliation{%
  \institution{AntGroup}
  \city{Hangzhou}
  \country{China}}
\email{mengshu.sms@antgroup.com}

\author{Zhiqiang Zhang}
\orcid{0000-0002-2321-7259}
\affiliation{%
  \institution{AntGroup}
  \city{Hangzhou}
  \country{China}}
\email{lingyao.zzq@antgroup.com}

\author{Jun Zhou}
\orcid{0000-0001-6033-6102}
\affiliation{%
  \institution{AntGroup}
  \city{Hangzhou}
  \country{China}}
\email{jun.zhoujun@antgroup.com}

\renewcommand{\shortauthors}{Xu et al.}

\begin{abstract}
Extracting event relations that deviate from known schemas has proven challenging for previous methods based on multi-class classification, MASK prediction, or prototype matching. Recent advancements in large language models have shown impressive performance through instruction tuning. Nevertheless, in the task of event relation extraction, instruction-based methods face several challenges: there are a vast number of inference samples, and the relations between events are non-sequential. To tackle these challenges, we present an improved instruction-based event relation extraction framework named MAQInstruct. Firstly, we transform the task from extracting event relations using given event-event instructions to selecting events using given event-relation instructions, which reduces the number of samples required for inference. Then, by incorporating a bipartite matching loss, we reduce the dependency of the instruction-based method on the generation sequence. Our experimental results demonstrate that MAQInstruct significantly improves the performance of event relation extraction across multiple LLMs.
\end{abstract}





\maketitle

\section{Introduction}
Event Relation Extraction (ERE) tasks
are highly diversified due to their varying sub-tasks (coreference, temporal, causal, sub-event, etc.) and complex relations (symmetrical, asymmetrical, cross, etc.)~\cite{DBLP:conf/emnlp/WenJ21,DBLP:conf/acl/WadhwaAW23,hu2023protoem,DBLP:conf/acl/Wang0GZC00LLXZL24}. 
However, most previous studies~\cite{DBLP:conf/emnlp/NguyenMDN22,DBLP:conf/eacl/WangZDGRC23,DBLP:conf/acl/YuanH0023,DBLP:conf/acl/0001MS0Z024,DBLP:conf/coling/XuSZZ24} have primarily focused on optimizing specific sub-tasks, making it difficult to transfer model structures, optimization strategies, specialized knowledge sources, and domain data between different sub-tasks. Although a few of works~\cite{wang-etal-2022-maven,hu2023protoem} use multi-head classification or prototype matching to tackle multiple sub-tasks simultaneously, these methods depend on pre-defined and mutually exclusive event relations.
\begin{figure}[htbp]
    \centering
    \includegraphics[width=8.5cm]{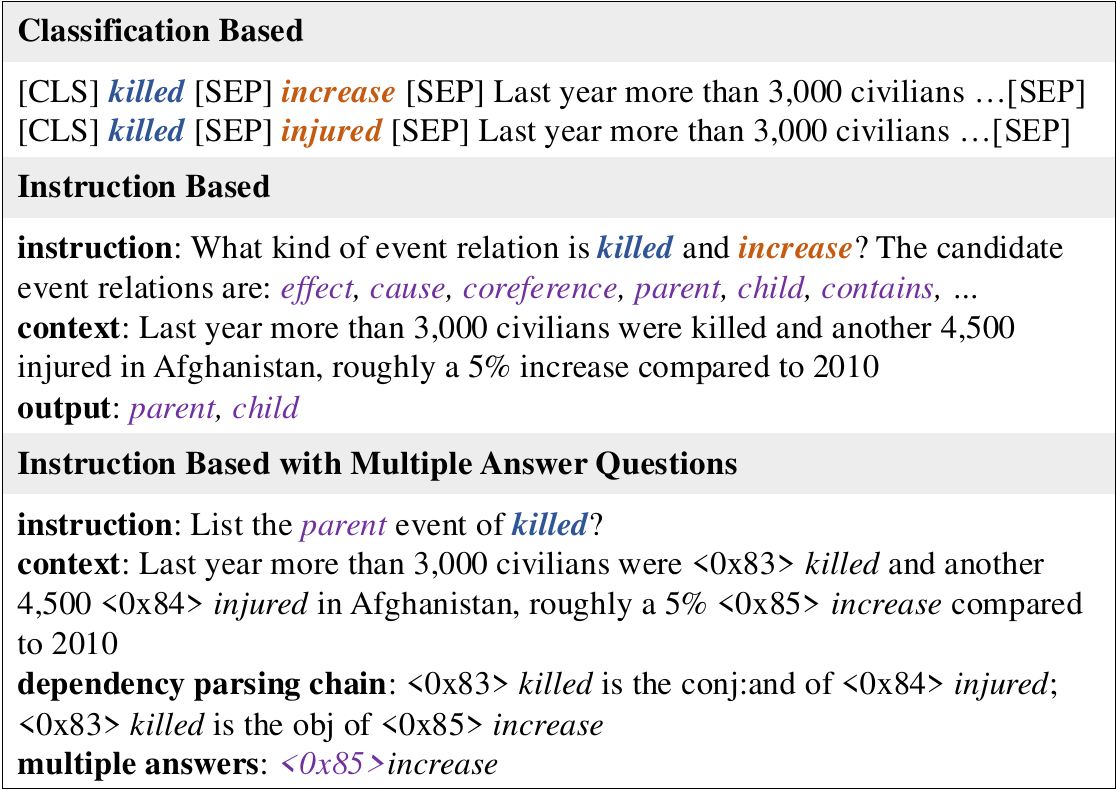}
    \caption{Different event relation extraction methods. 
    }
    \label{Fig.motivation}
\end{figure}
Recent large language models, such as ChatGPT and Llama, demonstrate exceptional text understanding and instruction-learning capabilities. The instruction-based approach eliminates the need for predefined relation schemas and mutual exclusivity in event relations, effectively solving previous issues. For a more intuitive comparison, we present the different methods in Figure~\ref{Fig.motivation}. 
The classification-based method utilizes one-hot embedding to represent the event relation labels, which overlook the semantic information of the labels and struggle to effectively extract unseen event relations. The instruction-based method uses pairs of candidate event mentions and a comprehensive list of all event relations for instructions, leveraging the capabilities of large language models to generate these relations. Although the instruction-based method can address the issues present in classification-based methods, it also has two significant drawbacks. 
First, it requires a large amount of training and inference samples, reaching $n \times n$, where $n$  represents the number of event mentions (dozens or hundreds within a single sample). 
Second, the model is greatly influenced by the sequence in which relations are generated. Using the instruction-based method shown in Figure~\ref{Fig.motivation} as an example, the model generates $p(parent|child)$ and $p(child|parent)$ with varying probabilities. However, in the ERE task, the sequence of generation should not affect the event relation between event mentions.

To address the two issues present in the instruction-based method, we design two strategies: multiple-answer question answering and bipartite matching. First, as shown in Figure~\ref{Fig.motivation}, we transform the task from extracting event relations using given event-event instructions to selecting events using given event-relation instructions. 
\begin{figure}[htbp]
\centering 
\includegraphics[width=8.5cm]{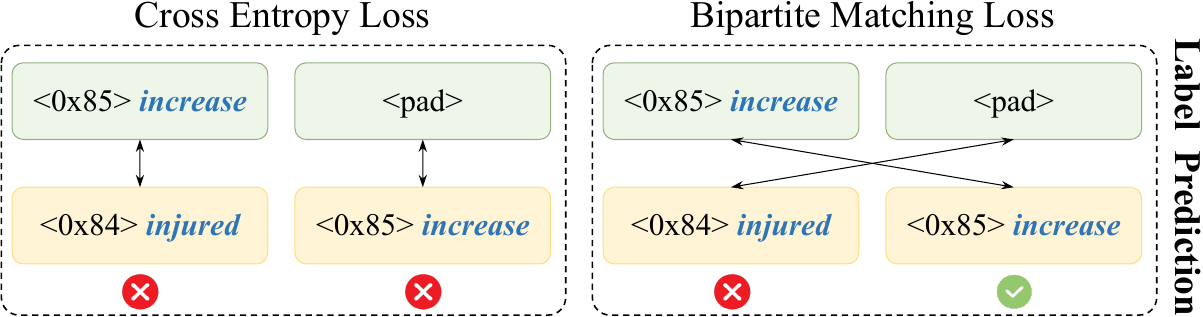} 
\caption{Cross-entropy loss vs bipartite matching loss.} 
\label{Fig.loss_function} 
\end{figure}
Since the event relation types 
$k\ll n$ (where $k$ has a few kinds, and $n$ is in the dozens or hundreds), we reduce the training and inference samples from $n\times n$ to $k \times n$. Second, based on the instruction-based framework, we introduce a bipartite matching loss. As demonstrated in Figure~\ref{Fig.loss_function}, using cross-entropy loss results in two mistakes, while the bipartite matching loss yields one correct answer and one mistake.
In summary, the main contributions of this paper are:

1) We propose an improved unified event relation extraction framework (MAQInstruct) based on multiple answer questions. Compared with InstructERE, our method reduces the training and inference samples from $n\times n$ to $k\times n$.

2) In the MAQInstruct framework, we incorporate a bipartite matching loss to reduce the dependency of InstructERE on the generation sequence, making it more suitable for event relation extraction tasks.

\section{Methodology}
\subsection{Sample Construction}
\noindent{\textbf{Instruction:}}
To unify the various inputs for different ERE sub-tasks, we have developed a set of instructions. Each instruction specifies an event relation and a candidate event mention, indicated by the special character <0x64>-<0xFF> in LLMs. For instance, the instruction ``List the coreference event of <0x85> ruled ?'' extracts coreferential events, with ``coreference'' as the relation and ``<0x85> ruled'' as the mention.

\noindent{\textbf{Context:}}
In the ERE task, we insert sequential markers (<0x64>-<0xFF>) into the text for each candidate event mention. The first mention receives <0x64>, the second <0x65>, and so on. These markers direct the language model to focus solely on the specified content. These markers enable LLMs to select the correct answer from these candidate events.

\noindent{\textbf{Label:}}
The output consists of two parts: the dependency parsing chain and multiple answers, separated by a colon. 
Multiple answers are presented in the order they appear in the text, separated by commas; if there are no associated event mentions, this part will be set to none.

\subsection{Multiple Answer Questions Loss}
The generated sequence greatly influences the text generation effectiveness~\citep{ye-etal-2021-one2set,OTSeq2SetCaoZ22}, but in ERE, the order of answer generation should not affect the outcome. To minimize the impact of the generation sequence, we calculate distinct losses for the dependency parsing chain and multiple answers, defined as follows:
\begin{equation}
\begin{aligned}
\mathcal{L}_{CE} = \frac{1}{N}\sum_{i=0}^{N}CE(y_i,p(y_i|x))
\end{aligned}
\label{sft_loss}
\end{equation}
where $N=N_1+N_2$, $N_1$ represents the length of dependency parsing chain and $N_2$ represents the length of multiple answers. CE is the cross-entropy loss. 
As illustrated in Figure~\ref{Fig.loss_function}, the sequence of generation does not impact the multiple answers. The loss for multiple answers is calculated as follows:

(a) First, use the Hungarian Algorithm to find the optimal match.
\begin{equation}
\begin{aligned}
\hat \theta = \mathop{\arg\min}\limits_{\theta \in \Psi_{N_2}} \sum_{i=0}^{N_2} 1-\log \hat p_{ \theta(i)}(c_i)
\end{aligned}
\label{ce_loss}
\end{equation}

(b) After optimal allocation, the loss function for Multiple Answers is:
\begin{equation}
\begin{aligned}
\mathcal{L}_{BPM}= \sum_{i=0}^{N_2} 1-\log \hat p_{\hat \theta(i)}(c_i)
\end{aligned}
\label{bpm_loss}
\end{equation}

(c) Finally, the total loss is as follows:
\begin{equation}
\begin{aligned}
\mathcal{L}= \mathcal{L}_{CE} + \lambda \mathcal{L}_{BPM}
\end{aligned}
\label{final_loss}
\end{equation}
where $\Psi_{N_2}$ denotes a permutation of $N_2$. $\theta$ is one of the permutations. $\theta(i)$ represents the i-th element in permutation $\theta$. $c_i$ represents the target vocabulary id of the i-th element. The probability of the i-th element in the permutation $\theta$ belonging to the target vocabulary id is denoted by $\hat p_{\theta(i)}(c_i)$. $\hat \theta$ stands for the optimal permutation. The weight parameter is represented by $\lambda$.

\renewcommand\arraystretch{1.0}
\begin{table*}[htbp]
\centering
\small
\caption{Performance of different models on the unified event relation extraction dataset MAVEN-ERE. Llama2 and GPT-4 indicate the results without training.}
\setlength\aboverulesep{0pt}\setlength\belowrulesep{0pt}
\begin{tabular}{p{2.9cm}|p{0.5cm}<{\centering}p{0.8cm}<{\centering}p{0.7cm}<{\centering}p{1.0cm}<{\centering}|p{0.75cm}<{\centering}p{0.75cm}<{\centering}p{0.75cm}<{\centering}|p{0.75cm}<{\centering}p{0.75cm}<{\centering}p{0.75cm}<{\centering}|p{0.75cm}<{\centering}p{0.75cm}<{\centering}p{0.75cm}<{\centering}}
\toprule
\multirow{2}{*}{Models} & \multicolumn{4}{c|}{COREFERENCE} & \multicolumn{3}{c|}{TEMPORAL} & \multicolumn{3}{c|}{CAUSAL} & \multicolumn{3}{c}{SUBEVENT} \\ \cline{2-14}
            & B$^3$ & CEAF$_e$ & MUC  & BLANC & P & R & F1   & P & R & F1   & P & R & F1   \\ \bottomrule
Llama2 & 43.5 & 35.7 & 22.1 & 33.5  & 12.6 & 11.1 & 11.8 & 7.5 & 10.9 & 8.9 & 8.3  & 5.4 & 6.5 \\ 
GPT-4 & 65.2 & 63.4 & 48.5 & 52.6 & 25.2 & 26.5 & 25.9 & 15.7 & 17.3 & 16.5 & 15.0 & 12.4 & 13.6 \\ 
BertERE & {97.8} & {97.6} & {79.8} & {88.3} & {50.9} & {53.4} & {52.1} & {31.3} & {30.5} & {30.9} & {24.6} & {22.9} & {23.7} \\ \bottomrule
\rowcolor{blue!10} InstructERE$_{\rm ChatGLM3}$  & 92.3 & 94.2 & 74.9 & 82.6 & 46.8 & 49.7 & 48.3 & 26.1 & 27.4 &26.7 & 19.2 & 20.1 & 19.6 \\ 
\rowcolor{red!10} InstructERE$_{\rm Qwen}$ & 93.5 & 94.6 & 74.1 & 85.2  & 48.0  & 51.6 & 49.3 & 27.8 & 28.1 & 27.9 & 20.3 & 21.3 & 20.8 \\ 
\rowcolor{green!10} InstructERE$_{\rm Llama2}$          & 94.2  & 93.5     & 73.3 & 84.7  & 48.5 & 51.0 & 49.7 & 28.6 & 28.0 & 28.3 & 20.9 & 21.7 & 21.3 \\ \bottomrule
\rowcolor{blue!10} MAQInstruct$_{\rm ChatGLM3}$ &  96.5 & 96.6 & 79.7 & 86.3 & 51.7 & 53.5 & 52.6 & 31.9 & 30.1 & 31.0 & 24.1 & 23.8 & 24.0 \\
\rowcolor{red!10} MAQInstruct$_{\rm Qwen}$ & 97.8 & 97.2 & 80.1 & 88.6 & 53.1 & 54.9 & \textbf{54.0} & 33.1 & 31.2 & 32.1 & 25.4  & 24.2 & 24.8 \\
\rowcolor{green!10} MAQInstruct$_{\rm Llama2}$ & \textbf{98.1} & \textbf{97.9} & \textbf{80.2} & \textbf{88.9} & {53.3} & {54.3} & 53.8 & {33.4} & {31.6} & \textbf{32.5} & {25.8} & {24.6} & \textbf{25.2} \\\bottomrule
\end{tabular}
\label{Tab.joint_result}
\end{table*}

\section{Experimental Results}
\subsection{Datasets \& Comparison Methods}
\noindent{\textbf{Dataset.}} Our experiments are conducted on four datasets, including MAVEN-ERE~\citep{wang-etal-2022-maven} for unified event relation extraction, HiEve~\citep{DBLP:conf/lrec/GlavasSMK14} for sub-event relation extraction, MATRES~\citep{DBLP:conf/acl/RothWN18} for temporal relation extraction, and MECI~\citep{DBLP:conf/coling/LaiVNDN22} for causal relation extraction.  

\noindent{\textbf{Comparison Methods.}}
\textbf{BertERE} encodes the entire document with RoBERTa, adding a classification head for contextualized representations of various event pairs, and fine-tunes the model for relation classification. \textbf{InstructERE} uses candidate event pairs and relations as instructions, utilizing a large language model to generate the correct event relations. 
\subsection{Overall Results}
The overall experimental results are summarized in Table~\ref{Tab.joint_result}. The performance of LLMs, specifically Llama2 and GPT-4, exhibits relative inadequacy in the ERE task when these models are not trained with instructions. This performance deficiency is attributed to the intricate definitions of event relations, which complicate comprehension for LLMs. Within a unified dataset MAVEN-ERE, BertERE demonstrates commendable performance, as there is no overlap between distinct event relations. While InstructERE leverages the excellent understanding capabilities of LLMs, it still does not surpass BertERE across various LLMs, indicating that InstructERE possesses inherent limitations in supervised event relation extraction tasks. Analyses indicate that InstructERE requires the construction of $n^2$ samples, which often contain considerable overlapping content, thereby hindering the learning process for LLMs. The model MAQInstruct, built on multiple-answer questions and bipartite matching, surpasses InstructERE when evaluated on three LLMs: ChatGLM3 (ChatGLM3-6b), Qwen (Qwen-7B-Chat), and Llama2 (Llama2-7B-Chat). Notably, the MAQInstruct model trained on Llama2 exhibits improvements of 4.9\%, 4.1\%, 4.2\%, and 3.9\% in event coreference, temporal, causal, and sub-event relations compared to InstructERE, respectively. Additionally, compared to BertERE, MAQInstruct shows enhancements of 0.4\%, 1.7\%, 1.6\%, and 1.5\%, respectively.

\subsection{Inference Performance Analysis}
The construction forms of samples for the three methods: BertERE, InstructERE, and MAQInstruct are shown in Figure~\ref{Fig.motivation}. As can be seen from Table~\ref{Tab.inference_performance_analysis}, the number of coreference event samples constructed by BertERE and InstructERE is significantly larger than that of MAQInstruct. Using an A100-80G GPU for inference with the same base model, Llama2, while keeping the batch size and sequence length, the inference time of InstructERE is 32.5 times that of MAQInstruct. Compared to BertERE, the inference time of MAQInstruct is still greater, since the parameter count of Llama2 is 70 times that of Bert.
\renewcommand\arraystretch{1.1}
\begin{table}[htbp]
\centering
\small
\caption{The inference samples and cost associated with the event coreference relation dataset in MAVEN-ERE.}
\setlength\aboverulesep{0pt}\setlength\belowrulesep{0pt}
\begin{tabular}{p{1.8cm}|p{1.2cm}<{\centering}|p{1.2cm}<{\centering}|p{1.2cm}<{\centering}|p{1.2cm}<{\centering}}
\toprule
            & \#doc & \#mention & \#query & cost(min) \\ \hline
BertERE     & 710   & 17,780    & 631,486 & 16        \\ \hline
InstructERE & 710   & 17,780    & 631,486 & 813       \\ \hline
MAQInstruct & 710   & 17,780    & 17,780  & 25       \\ \bottomrule
\end{tabular}
\label{Tab.inference_performance_analysis}
\end{table}

\subsection{Bipartite Matching Loss Analysis}
\begin{figure}[htbp]
\centering 
\includegraphics[width=8.3cm]{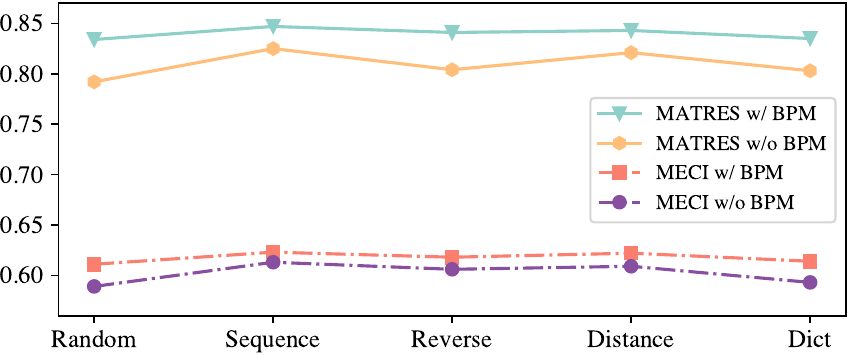} 
\caption{The performance of different answer sequences.} 
\label{Fig.bpm_analysis} 
\end{figure}

The performance of a generative model is greatly affected by the generation sequence, as shown in Figure~\ref{Fig.bpm_analysis}. "Random" indicates that the answers are in a random sequence, "Sequence" represents the sequence in which they appear in the text, "Reverse" indicates the reverse sequence of their appearance, "Distance" means the answers are sorted by distance from the query mention, and "Dict" sorts them from A to Z.
When the bipartite matching loss is not considered, random answer sequences perform the worst, with a reduction of 4.00\% and 3.92\% compared to ordered sequences in MATRES and MECI, respectively. However, after incorporating the bipartite matching loss, MAQInstruct is capable of effectively generating the correct results with any answer sequence used.

\subsection{Zero-Shot Learning Analysis}
\begin{figure}[htbp]
\centering 
\includegraphics[width=5cm]{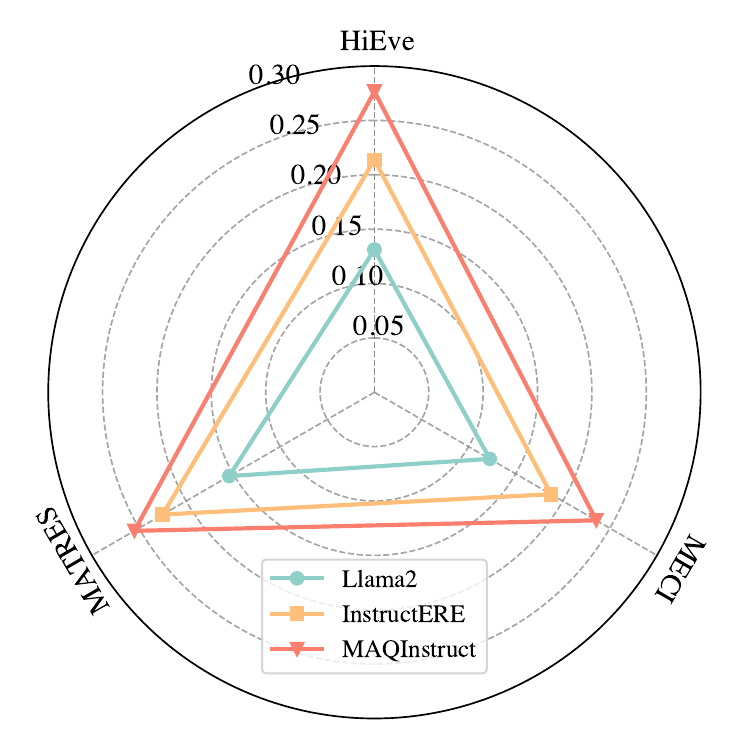} 
\caption{The F1 score of different instruction-based LLMs on the zero-shot event relation extraction task.} 
\label{Fig.zeroshot_ere_analysis} 
\end{figure}

\begin{figure}[htbp]
\centering 
\includegraphics[width=8.3cm]{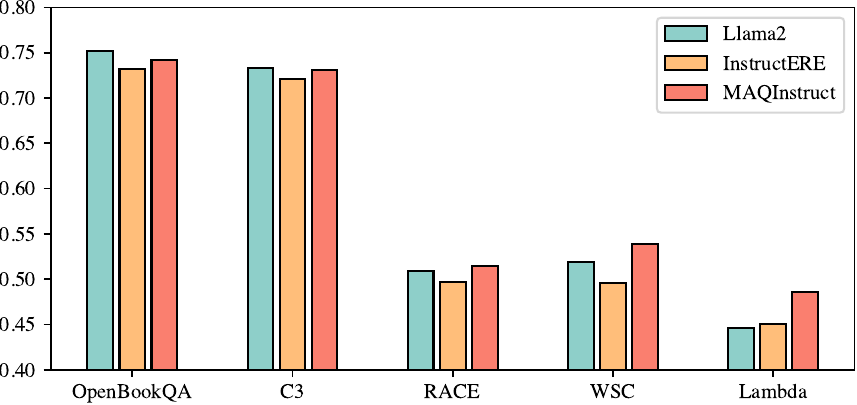} 
\caption{The performance of different LLMs on natural language understanding evaluation datasets.} 
\label{Fig.text_understanding_analysis} 
\end{figure}

To validate the zero-shot learning capability of the model, experiments are conducted on three event relation extraction datasets: HiEve, MATRES, and MECI. 
The experimental results are presented in Figure~\ref{Fig.zeroshot_ere_analysis}. 
For instance, when comparing InstructERE and MAQInstruct, both trained with Llama2, it is observed that they significantly outperform the untrained Llama2 in event relation extraction tasks. 
The primary reason for this enhancement lies in the fact that LLMs trained with event relation extraction instructions possess a more comprehensive understanding of the definitions of event relations. 
Furthermore, the performance of MAQInstruct surpasses that of InstructERE, indicating that instruction based on multiple-answer questions is more effective for event relation extraction.
Additionally, to evaluate the effectiveness of data constructed from multiple-answer questions, the performance of MAQInstruct in general natural language understanding tasks is also assessed. As illustrated in Figure~\ref{Fig.text_understanding_analysis}, the average performance of MAQInstruct slightly exceeds that of InstructERE and Llama2, demonstrating that neither the method of data construction from multiple answer questions nor the bipartite matching training approach adversely impacts the language understanding performance of LLMs.

\subsection{Model Ablation Studies}
Due to the substantial volume of MAVEN-ERE data, conducting ablation experiments proves to be excessively costly; consequently, we opt to perform ablation on MATRES and MECI. The experimental results are presented in Table~\ref{Tab.ablation}. 
\renewcommand\arraystretch{1.1}
\begin{table}[htbp]
\centering
\small
\caption{Model ablation studies. Marker refers to the identifier that precedes a  event mention, e.g., ``<0x8F>''. DPC means dependency parsing chain.}
\setlength\aboverulesep{0pt}\setlength\belowrulesep{0pt}
\begin{tabular}{p{1.7cm}|p{0.7cm}<{\centering} p{0.7cm}<{\centering} p{0.7cm}<{\centering}|p{0.7cm}<{\centering} p{0.7cm}<{\centering} p{0.7cm}<{\centering}}
\toprule
\multirow{2}{*}{Method} & \multicolumn{3}{c|}{MATRES} & \multicolumn{3}{c}{MECI} \\ \cline{2-7}
                       & P       & R       & F1     & P      & R      & F1     \\ \hline
MAQInstruct            & 85.5    & 83.9    & 84.7   & 62.9   & 61.6   & 62.3   \\ \hline
w/o Marker             & 81.2    & 84.1    & 82.6   & 58.9   & 59.3   & 59.1   \\ \hline
only Marker            & 84.6    & 83.8    & 84.2   & 62.2   & 61.8   & 62.0   \\ \hline
w/o DPC                & 82.3    & 80.5    & 81.4   & 57.5   & 59.3   & 58.4   \\ \hline
w/o $\mathcal{L}_{BPM}$& 81.4    & 83.6    & 82.5   & 61.1   & 61.5   & 61.3   \\ 
\bottomrule
\end{tabular}
\label{Tab.ablation}
\end{table}
Initially, in the absence of the marker (<0x64>-<0xFF>), we observe performance declines of 2.48\% on MATRES and 5.14\% on MECI, which substantiates the efficacy of the prefix marker. In scenarios where multiple answers consist solely of markers, such as ``<0x84>, <0x85>'' instead of ``<0x84> injured, <0x85> increase'', this results in a marginal decrease in effectiveness, suggesting that these markers may lack complete semantic information. 
Moreover, the elimination of the dependency parsing chain results in the most significant performance decline. This phenomenon can be attributed to the fact that the dependency parsing chain enhances the model's capability to extract scattered event relations by utilizing structured information. The removal of the bipartite matching loss function causes a significant drop in model effectiveness, indicating that the bipartite matching loss is particularly suitable for scenarios in which the sequence of generated results is not predetermined.

\section{Conclusion}
In this study, we present a unified framework called MAQInstruct, which aims to improve instruction-based methods through multiple-answer questions, effectively extracting various event relations via different types of instructions. By utilizing the instruction-based method, MAQInstruct significantly enhances this model’s performance by introducing strategies such as multiple-answer questions and bipartite matching loss. 
\bibliographystyle{ACM-Reference-Format}
\bibliography{sample-base}


\begin{thebibliography}{43}


\ifx \showCODEN    \undefined \def \showCODEN     #1{\unskip}     \fi
\ifx \showISBNx    \undefined \def \showISBNx     #1{\unskip}     \fi
\ifx \showISBNxiii \undefined \def \showISBNxiii  #1{\unskip}     \fi
\ifx \showISSN     \undefined \def \showISSN      #1{\unskip}     \fi
\ifx \showLCCN     \undefined \def \showLCCN      #1{\unskip}     \fi
\ifx \shownote     \undefined \def \shownote      #1{#1}          \fi
\ifx \showarticletitle \undefined \def \showarticletitle #1{#1}   \fi
\ifx \showURL      \undefined \def \showURL       {\relax}        \fi
\providecommand\bibfield[2]{#2}
\providecommand\bibinfo[2]{#2}
\providecommand\natexlab[1]{#1}
\providecommand\showeprint[2][]{arXiv:#2}

\bibitem[Bagga and Baldwin(1998)]%
        {bagga1998algorithms}
\bibfield{author}{\bibinfo{person}{Amit Bagga} {and} \bibinfo{person}{Breck Baldwin}.} \bibinfo{year}{1998}\natexlab{}.
\newblock \showarticletitle{Algorithms for scoring coreference chains}. In \bibinfo{booktitle}{\emph{The first international conference on language resources and evaluation workshop on linguistics coreference}}, Vol.~\bibinfo{volume}{1}. \bibinfo{pages}{563--566}.
\newblock


\bibitem[Barhom et~al\mbox{.}(2019)]%
        {DBLP:conf/acl/BarhomSEBRD19}
\bibfield{author}{\bibinfo{person}{Shany Barhom}, \bibinfo{person}{Vered Shwartz}, \bibinfo{person}{Alon Eirew}, \bibinfo{person}{Michael Bugert}, \bibinfo{person}{Nils Reimers}, {and} \bibinfo{person}{Ido Dagan}.} \bibinfo{year}{2019}\natexlab{}.
\newblock \showarticletitle{Revisiting Joint Modeling of Cross-document Entity and Event Coreference Resolution}. In \bibinfo{booktitle}{\emph{ACL}}. \bibinfo{pages}{4179--4189}.
\newblock


\bibitem[Cao and Zhang(2022)]%
        {OTSeq2SetCaoZ22}
\bibfield{author}{\bibinfo{person}{Jie Cao} {and} \bibinfo{person}{Yin Zhang}.} \bibinfo{year}{2022}\natexlab{}.
\newblock \showarticletitle{OTSeq2Set: An Optimal Transport Enhanced Sequence-to-Set Model for Extreme Multi-label Text Classification}. In \bibinfo{booktitle}{\emph{{EMNLP}}}. \bibinfo{pages}{5588--5597}.
\newblock


\bibitem[Caselli and Vossen(2017)]%
        {DBLP:conf/acl/CaselliV17}
\bibfield{author}{\bibinfo{person}{Tommaso Caselli} {and} \bibinfo{person}{Piek Vossen}.} \bibinfo{year}{2017}\natexlab{}.
\newblock \showarticletitle{The Event StoryLine Corpus: {A} New Benchmark for Causal and Temporal Relation Extraction}. In \bibinfo{booktitle}{\emph{ACL}}. \bibinfo{pages}{77--86}.
\newblock


\bibitem[Chen et~al\mbox{.}(2024)]%
        {DBLP:conf/acl/0001MS0Z024}
\bibfield{author}{\bibinfo{person}{Meiqi Chen}, \bibinfo{person}{Yubo Ma}, \bibinfo{person}{Kaitao Song}, \bibinfo{person}{Yixin Cao}, \bibinfo{person}{Yan Zhang}, {and} \bibinfo{person}{Dongsheng Li}.} \bibinfo{year}{2024}\natexlab{}.
\newblock \showarticletitle{Improving Large Language Models in Event Relation Logical Prediction}. In \bibinfo{booktitle}{\emph{ACL}}. \bibinfo{pages}{9451--9478}.
\newblock


\bibitem[Choubey and Huang(2017)]%
        {DBLP:conf/emnlp/ChoubeyH17a}
\bibfield{author}{\bibinfo{person}{Prafulla~Kumar Choubey} {and} \bibinfo{person}{Ruihong Huang}.} \bibinfo{year}{2017}\natexlab{}.
\newblock \showarticletitle{Event Coreference Resolution by Iteratively Unfolding Inter-dependencies among Events}. In \bibinfo{booktitle}{\emph{EMNLP}}. \bibinfo{pages}{2124--2133}.
\newblock


\bibitem[Cui et~al\mbox{.}(2022)]%
        {DBLP:conf/coling/CuiSCLLS22}
\bibfield{author}{\bibinfo{person}{Shiyao Cui}, \bibinfo{person}{Jiawei Sheng}, \bibinfo{person}{Xin Cong}, \bibinfo{person}{Quangang Li}, \bibinfo{person}{Tingwen Liu}, {and} \bibinfo{person}{Jinqiao Shi}.} \bibinfo{year}{2022}\natexlab{}.
\newblock \showarticletitle{Event Causality Extraction with Event Argument Correlations}. In \bibinfo{booktitle}{\emph{COLING}}. \bibinfo{pages}{2300--2312}.
\newblock


\bibitem[Glavas et~al\mbox{.}(2014)]%
        {DBLP:conf/lrec/GlavasSMK14}
\bibfield{author}{\bibinfo{person}{Goran Glavas}, \bibinfo{person}{Jan Snajder}, \bibinfo{person}{Marie{-}Francine Moens}, {and} \bibinfo{person}{Parisa Kordjamshidi}.} \bibinfo{year}{2014}\natexlab{}.
\newblock \showarticletitle{HiEve: {A} Corpus for Extracting Event Hierarchies from News Stories}. In \bibinfo{booktitle}{\emph{LREC}}. \bibinfo{pages}{3678--3683}.
\newblock


\bibitem[Hu et~al\mbox{.}(2019)]%
        {hu-etal-2019-multi}
\bibfield{author}{\bibinfo{person}{Minghao Hu}, \bibinfo{person}{Yuxing Peng}, \bibinfo{person}{Zhen Huang}, {and} \bibinfo{person}{Dongsheng Li}.} \bibinfo{year}{2019}\natexlab{}.
\newblock \showarticletitle{A Multi-Type Multi-Span Network for Reading Comprehension that Requires Discrete Reasoning}. In \bibinfo{booktitle}{\emph{EMNLP-IJCNLP}}. \bibinfo{address}{Hong Kong, China}.
\newblock


\bibitem[Hu et~al\mbox{.}(2023a)]%
        {DBLP:conf/acl/HuLJ0GGC23}
\bibfield{author}{\bibinfo{person}{Zhilei Hu}, \bibinfo{person}{Zixuan Li}, \bibinfo{person}{Xiaolong Jin}, \bibinfo{person}{Long Bai}, \bibinfo{person}{Saiping Guan}, \bibinfo{person}{Jiafeng Guo}, {and} \bibinfo{person}{Xueqi Cheng}.} \bibinfo{year}{2023}\natexlab{a}.
\newblock \showarticletitle{Semantic Structure Enhanced Event Causality Identification}. In \bibinfo{booktitle}{\emph{ACL}}. \bibinfo{pages}{10901--10913}.
\newblock


\bibitem[Hu et~al\mbox{.}(2023b)]%
        {hu2023protoem}
\bibfield{author}{\bibinfo{person}{Zhilei Hu}, \bibinfo{person}{Zixuan Li}, \bibinfo{person}{Daozhu Xu}, \bibinfo{person}{Long Bai}, \bibinfo{person}{Cheng Jin}, \bibinfo{person}{Xiaolong Jin}, \bibinfo{person}{Jiafeng Guo}, {and} \bibinfo{person}{Xueqi Cheng}.} \bibinfo{year}{2023}\natexlab{b}.
\newblock \bibinfo{title}{ProtoEM: A Prototype-Enhanced Matching Framework for Event Relation Extraction}.
\newblock
\showeprint[arxiv]{2309.12892}~[cs.CL]


\bibitem[Huang et~al\mbox{.}(2023)]%
        {huang2023classification}
\bibfield{author}{\bibinfo{person}{Quzhe Huang}, \bibinfo{person}{Yutong Hu}, \bibinfo{person}{Shengqi Zhu}, \bibinfo{person}{Yansong Feng}, \bibinfo{person}{Chang Liu}, {and} \bibinfo{person}{Dongyan Zhao}.} \bibinfo{year}{2023}\natexlab{}.
\newblock \showarticletitle{More than Classification: {A} Unified Framework for Event Temporal Relation Extraction}. In \bibinfo{booktitle}{\emph{ACL}}. \bibinfo{pages}{9631--9646}.
\newblock


\bibitem[Hwang et~al\mbox{.}(2022)]%
        {hwang-etal-2022-event}
\bibfield{author}{\bibinfo{person}{EunJeong Hwang}, \bibinfo{person}{Jay-Yoon Lee}, \bibinfo{person}{Tianyi Yang}, \bibinfo{person}{Dhruvesh Patel}, \bibinfo{person}{Dongxu Zhang}, {and} \bibinfo{person}{Andrew McCallum}.} \bibinfo{year}{2022}\natexlab{}.
\newblock \showarticletitle{Event-Event Relation Extraction using Probabilistic Box Embedding}. In \bibinfo{booktitle}{\emph{ACL}}. \bibinfo{pages}{235--244}.
\newblock


\bibitem[Lai et~al\mbox{.}(2022)]%
        {DBLP:conf/coling/LaiVNDN22}
\bibfield{author}{\bibinfo{person}{Viet~Dac Lai}, \bibinfo{person}{Amir Pouran~Ben Veyseh}, \bibinfo{person}{Minh~Van Nguyen}, \bibinfo{person}{Franck Dernoncourt}, {and} \bibinfo{person}{Thien~Huu Nguyen}.} \bibinfo{year}{2022}\natexlab{}.
\newblock \showarticletitle{{MECI:} {A} Multilingual Dataset for Event Causality Identification}. In \bibinfo{booktitle}{\emph{COLING}}. \bibinfo{pages}{2346--2356}.
\newblock


\bibitem[Lu and Ng(2021)]%
        {DBLP:conf/naacl/LuN21}
\bibfield{author}{\bibinfo{person}{Jing Lu} {and} \bibinfo{person}{Vincent Ng}.} \bibinfo{year}{2021}\natexlab{}.
\newblock \showarticletitle{Constrained Multi-Task Learning for Event Coreference Resolution}. In \bibinfo{booktitle}{\emph{{NAACL-HLT}}}. \bibinfo{pages}{4504--4514}.
\newblock


\bibitem[Lu et~al\mbox{.}(2022)]%
        {lu-etal-2022-unified}
\bibfield{author}{\bibinfo{person}{Yaojie Lu}, \bibinfo{person}{Qing Liu}, \bibinfo{person}{Dai Dai}, \bibinfo{person}{Xinyan Xiao}, \bibinfo{person}{Hongyu Lin}, \bibinfo{person}{Xianpei Han}, \bibinfo{person}{Le Sun}, {and} \bibinfo{person}{Hua Wu}.} \bibinfo{year}{2022}\natexlab{}.
\newblock \showarticletitle{Unified Structure Generation for Universal Information Extraction}. In \bibinfo{booktitle}{\emph{ACL}}. \bibinfo{address}{Dublin, Ireland}, \bibinfo{pages}{5755--5772}.
\newblock


\bibitem[Luo(2005)]%
        {Xiaoqiang1220579Luo}
\bibfield{author}{\bibinfo{person}{Xiaoqiang Luo}.} \bibinfo{year}{2005}\natexlab{}.
\newblock \showarticletitle{On coreference resolution performance metrics}. In \bibinfo{booktitle}{\emph{HLT}} (Vancouver, British Columbia, Canada). \bibinfo{address}{Morristown, NJ, USA}, \bibinfo{pages}{25--32}.
\newblock


\bibitem[Man et~al\mbox{.}(2022)]%
        {hieu2022selecting}
\bibfield{author}{\bibinfo{person}{Hieu Man}, \bibinfo{person}{Nghia~Trung Ngo}, \bibinfo{person}{Linh~Ngo Van}, {and} \bibinfo{person}{Thien~Huu Nguyen}.} \bibinfo{year}{2022}\natexlab{}.
\newblock \showarticletitle{Selecting Optimal Context Sentences for Event-Event Relation Extraction}. In \bibinfo{booktitle}{\emph{AAAI}}. \bibinfo{pages}{11058--11066}.
\newblock


\bibitem[Nguyen et~al\mbox{.}(2022a)]%
        {DBLP:conf/emnlp/NguyenMDN22}
\bibfield{author}{\bibinfo{person}{Minh~Van Nguyen}, \bibinfo{person}{Bonan Min}, \bibinfo{person}{Franck Dernoncourt}, {and} \bibinfo{person}{Thien Nguyen}.} \bibinfo{year}{2022}\natexlab{a}.
\newblock \showarticletitle{Learning Cross-Task Dependencies for Joint Extraction of Entities, Events, Event Arguments, and Relations}. In \bibinfo{booktitle}{\emph{{EMNLP}}}. \bibinfo{pages}{9349--9360}.
\newblock


\bibitem[Nguyen et~al\mbox{.}(2022b)]%
        {DBLP:conf/naacl/NguyenMDN22}
\bibfield{author}{\bibinfo{person}{Minh~Van Nguyen}, \bibinfo{person}{Bonan Min}, \bibinfo{person}{Franck Dernoncourt}, {and} \bibinfo{person}{Thien~Huu Nguyen}.} \bibinfo{year}{2022}\natexlab{b}.
\newblock \showarticletitle{Joint Extraction of Entities, Relations, and Events via Modeling Inter-Instance and Inter-Label Dependencies}. In \bibinfo{booktitle}{\emph{{NAACL}}}. \bibinfo{pages}{4363--4374}.
\newblock


\bibitem[Ning et~al\mbox{.}(2018)]%
        {DBLP:conf/acl/RothWN18}
\bibfield{author}{\bibinfo{person}{Qiang Ning}, \bibinfo{person}{Hao Wu}, {and} \bibinfo{person}{Dan Roth}.} \bibinfo{year}{2018}\natexlab{}.
\newblock \showarticletitle{A Multi-Axis Annotation Scheme for Event Temporal Relations}. In \bibinfo{booktitle}{\emph{ACL}}. \bibinfo{pages}{1318--1328}.
\newblock


\bibitem[RECASENS and HOVY(2011)]%
        {recasens_hovy_2011}
\bibfield{author}{\bibinfo{person}{M. RECASENS} {and} \bibinfo{person}{E. HOVY}.} \bibinfo{year}{2011}\natexlab{}.
\newblock \showarticletitle{BLANC: Implementing the Rand index for coreference evaluation}.
\newblock \bibinfo{journal}{\emph{Natural Language Engineering}} \bibinfo{volume}{17}, \bibinfo{number}{4} (\bibinfo{year}{2011}), \bibinfo{pages}{485–510}.
\newblock


\bibitem[Segal et~al\mbox{.}(2020)]%
        {DBLP:conf/emnlp/SegalESGB20}
\bibfield{author}{\bibinfo{person}{Elad Segal}, \bibinfo{person}{Avia Efrat}, \bibinfo{person}{Mor Shoham}, \bibinfo{person}{Amir Globerson}, {and} \bibinfo{person}{Jonathan Berant}.} \bibinfo{year}{2020}\natexlab{}.
\newblock \showarticletitle{A Simple and Effective Model for Answering Multi-span Questions}. In \bibinfo{booktitle}{\emph{EMNLP}}. \bibinfo{pages}{3074--3080}.
\newblock


\bibitem[Shen et~al\mbox{.}(2022)]%
        {DBLP:conf/coling/ShenZWQ22}
\bibfield{author}{\bibinfo{person}{Shirong Shen}, \bibinfo{person}{Heng Zhou}, \bibinfo{person}{Tongtong Wu}, {and} \bibinfo{person}{Guilin Qi}.} \bibinfo{year}{2022}\natexlab{}.
\newblock \showarticletitle{Event Causality Identification via Derivative Prompt Joint Learning}. In \bibinfo{booktitle}{\emph{{COLING}}}. \bibinfo{pages}{2288--2299}.
\newblock


\bibitem[Tan et~al\mbox{.}(2023)]%
        {DBLP:conf/eacl/TanPH23}
\bibfield{author}{\bibinfo{person}{Xingwei Tan}, \bibinfo{person}{Gabriele Pergola}, {and} \bibinfo{person}{Yulan He}.} \bibinfo{year}{2023}\natexlab{}.
\newblock \showarticletitle{Event Temporal Relation Extraction with Bayesian Translational Model}. In \bibinfo{booktitle}{\emph{EACL}}. \bibinfo{pages}{1117--1130}.
\newblock


\bibitem[Tran et~al\mbox{.}(2021)]%
        {DBLP:conf/acl/TranPN20}
\bibfield{author}{\bibinfo{person}{Hieu~Minh Tran}, \bibinfo{person}{Duy Phung}, {and} \bibinfo{person}{Thien~Huu Nguyen}.} \bibinfo{year}{2021}\natexlab{}.
\newblock \showarticletitle{Exploiting Document Structures and Cluster Consistencies for Event Coreference Resolution}. In \bibinfo{booktitle}{\emph{{ACL/IJCNLP}}}. \bibinfo{pages}{4840--4850}.
\newblock


\bibitem[Vilain et~al\mbox{.}(1995)]%
        {Vilain1995AMC}
\bibfield{author}{\bibinfo{person}{Marc~B. Vilain}, \bibinfo{person}{John~D. Burger}, \bibinfo{person}{John~S. Aberdeen}, \bibinfo{person}{Dennis Connolly}, {and} \bibinfo{person}{Lynette Hirschman}.} \bibinfo{year}{1995}\natexlab{}.
\newblock \showarticletitle{A Model-Theoretic Coreference Scoring Scheme}. In \bibinfo{booktitle}{\emph{Message Understanding Conference}}.
\newblock


\bibitem[Wadhwa et~al\mbox{.}(2023)]%
        {DBLP:conf/acl/WadhwaAW23}
\bibfield{author}{\bibinfo{person}{Somin Wadhwa}, \bibinfo{person}{Silvio Amir}, {and} \bibinfo{person}{Byron~C. Wallace}.} \bibinfo{year}{2023}\natexlab{}.
\newblock \showarticletitle{Revisiting Relation Extraction in the era of Large Language Models}. In \bibinfo{booktitle}{\emph{ACL}}. \bibinfo{pages}{15566--15589}.
\newblock


\bibitem[Wang et~al\mbox{.}(2020)]%
        {DBLP:conf/emnlp/WangCZR20}
\bibfield{author}{\bibinfo{person}{Haoyu Wang}, \bibinfo{person}{Muhao Chen}, \bibinfo{person}{Hongming Zhang}, {and} \bibinfo{person}{Dan Roth}.} \bibinfo{year}{2020}\natexlab{}.
\newblock \showarticletitle{Joint Constrained Learning for Event-Event Relation Extraction}. In \bibinfo{booktitle}{\emph{{EMNLP}}}. \bibinfo{pages}{696--706}.
\newblock


\bibitem[Wang et~al\mbox{.}(2023a)]%
        {DBLP:conf/eacl/WangZDGRC23}
\bibfield{author}{\bibinfo{person}{Haoyu Wang}, \bibinfo{person}{Hongming Zhang}, \bibinfo{person}{Yuqian Deng}, \bibinfo{person}{Jacob~R. Gardner}, \bibinfo{person}{Dan Roth}, {and} \bibinfo{person}{Muhao Chen}.} \bibinfo{year}{2023}\natexlab{a}.
\newblock \showarticletitle{Extracting or Guessing? Improving Faithfulness of Event Temporal Relation Extraction}. In \bibinfo{booktitle}{\emph{EACL}}. \bibinfo{pages}{541--553}.
\newblock


\bibitem[Wang et~al\mbox{.}(2022)]%
        {wang-etal-2022-maven}
\bibfield{author}{\bibinfo{person}{Xiaozhi Wang}, \bibinfo{person}{Yulin Chen}, \bibinfo{person}{Ning Ding}, \bibinfo{person}{Hao Peng}, \bibinfo{person}{Zimu Wang}, \bibinfo{person}{Yankai Lin}, \bibinfo{person}{Xu Han}, \bibinfo{person}{Lei Hou}, \bibinfo{person}{Juanzi Li}, \bibinfo{person}{Zhiyuan Liu}, \bibinfo{person}{Peng Li}, {and} \bibinfo{person}{Jie Zhou}.} \bibinfo{year}{2022}\natexlab{}.
\newblock \showarticletitle{{MAVEN}-{ERE}: A Unified Large-scale Dataset for Event Coreference, Temporal, Causal, and Subevent Relation Extraction}. In \bibinfo{booktitle}{\emph{{EMNLP}}}. \bibinfo{pages}{926--941}.
\newblock


\bibitem[Wang et~al\mbox{.}(2024)]%
        {DBLP:conf/acl/Wang0GZC00LLXZL24}
\bibfield{author}{\bibinfo{person}{Xiaozhi Wang}, \bibinfo{person}{Hao Peng}, \bibinfo{person}{Yong Guan}, \bibinfo{person}{Kaisheng Zeng}, \bibinfo{person}{Jianhui Chen}, \bibinfo{person}{Lei Hou}, \bibinfo{person}{Xu Han}, \bibinfo{person}{Yankai Lin}, \bibinfo{person}{Zhiyuan Liu}, \bibinfo{person}{Ruobing Xie}, \bibinfo{person}{Jie Zhou}, {and} \bibinfo{person}{Juanzi Li}.} \bibinfo{year}{2024}\natexlab{}.
\newblock \showarticletitle{{MAVEN-ARG:} Completing the Puzzle of All-in-One Event Understanding Dataset with Event Argument Annotation}. In \bibinfo{booktitle}{\emph{ACL}}. \bibinfo{pages}{4072--4091}.
\newblock


\bibitem[Wang et~al\mbox{.}(2023b)]%
        {wang2023instructuie}
\bibfield{author}{\bibinfo{person}{Xiao Wang}, \bibinfo{person}{Weikang Zhou}, \bibinfo{person}{Can Zu}, \bibinfo{person}{Han Xia}, \bibinfo{person}{Tianze Chen}, \bibinfo{person}{Yuansen Zhang}, \bibinfo{person}{Rui Zheng}, \bibinfo{person}{Junjie Ye}, \bibinfo{person}{Qi Zhang}, \bibinfo{person}{Tao Gui}, \bibinfo{person}{Jihua Kang}, \bibinfo{person}{Jingsheng Yang}, \bibinfo{person}{Siyuan Li}, {and} \bibinfo{person}{Chunsai Du}.} \bibinfo{year}{2023}\natexlab{b}.
\newblock \bibinfo{title}{InstructUIE: Multi-task Instruction Tuning for Unified Information Extraction}.
\newblock
\showeprint[arxiv]{2304.08085}~[cs.CL]


\bibitem[Wen and Ji(2021)]%
        {DBLP:conf/emnlp/WenJ21}
\bibfield{author}{\bibinfo{person}{Haoyang Wen} {and} \bibinfo{person}{Heng Ji}.} \bibinfo{year}{2021}\natexlab{}.
\newblock \showarticletitle{Utilizing Relative Event Time to Enhance Event-Event Temporal Relation Extraction}. In \bibinfo{booktitle}{\emph{{EMNLP}}}. \bibinfo{pages}{10431--10437}.
\newblock


\bibitem[Xiang et~al\mbox{.}(2023)]%
        {DBLP:journals/corr/abs-2307-09813}
\bibfield{author}{\bibinfo{person}{Wei Xiang}, \bibinfo{person}{Chuanhong Zhan}, {and} \bibinfo{person}{Bang Wang}.} \bibinfo{year}{2023}\natexlab{}.
\newblock \showarticletitle{DAPrompt: Deterministic Assumption Prompt Learning for Event Causality Identification}.
\newblock \bibinfo{journal}{\emph{CoRR}}  \bibinfo{volume}{abs/2307.09813} (\bibinfo{year}{2023}).
\newblock


\bibitem[Xiao et~al\mbox{.}(2024)]%
        {xiao2024yayiuie}
\bibfield{author}{\bibinfo{person}{Xinglin Xiao}, \bibinfo{person}{Yijie Wang}, \bibinfo{person}{Nan Xu}, \bibinfo{person}{Yuqi Wang}, \bibinfo{person}{Hanxuan Yang}, \bibinfo{person}{Minzheng Wang}, \bibinfo{person}{Yin Luo}, \bibinfo{person}{Lei Wang}, \bibinfo{person}{Wenji Mao}, {and} \bibinfo{person}{Daniel Zeng}.} \bibinfo{year}{2024}\natexlab{}.
\newblock \bibinfo{title}{YAYI-UIE: A Chat-Enhanced Instruction Tuning Framework for Universal Information Extraction}.
\newblock
\showeprint[arxiv]{2312.15548}~[cs.CL]


\bibitem[Xu et~al\mbox{.}(2024)]%
        {DBLP:conf/coling/XuSZZ24}
\bibfield{author}{\bibinfo{person}{Jun Xu}, \bibinfo{person}{Mengshu Sun}, \bibinfo{person}{Zhiqiang Zhang}, {and} \bibinfo{person}{Jun Zhou}.} \bibinfo{year}{2024}\natexlab{}.
\newblock \showarticletitle{ChatUIE: Exploring Chat-based Unified Information Extraction Using Large Language Models}. In \bibinfo{booktitle}{\emph{{LREC/COLING} 2024}}. \bibinfo{pages}{3146--3152}.
\newblock


\bibitem[Xu et~al\mbox{.}(2022)]%
        {xu-etal-2022-extracting}
\bibfield{author}{\bibinfo{person}{Jun Xu}, \bibinfo{person}{Weidi Xu}, \bibinfo{person}{Mengshu Sun}, \bibinfo{person}{Taifeng Wang}, {and} \bibinfo{person}{Wei Chu}.} \bibinfo{year}{2022}\natexlab{}.
\newblock \showarticletitle{Extracting Trigger-sharing Events via an Event Matrix}. In \bibinfo{booktitle}{\emph{EMNLP}}. \bibinfo{address}{Abu Dhabi, United Arab Emirates}, \bibinfo{pages}{1189--1201}.
\newblock


\bibitem[Ye et~al\mbox{.}(2022)]%
        {ye-etal-2022-packed}
\bibfield{author}{\bibinfo{person}{Deming Ye}, \bibinfo{person}{Yankai Lin}, \bibinfo{person}{Peng Li}, {and} \bibinfo{person}{Maosong Sun}.} \bibinfo{year}{2022}\natexlab{}.
\newblock \showarticletitle{Packed Levitated Marker for Entity and Relation Extraction}. In \bibinfo{booktitle}{\emph{ACL}}. \bibinfo{pages}{4904--4917}.
\newblock


\bibitem[Ye et~al\mbox{.}(2021)]%
        {ye-etal-2021-one2set}
\bibfield{author}{\bibinfo{person}{Jiacheng Ye}, \bibinfo{person}{Tao Gui}, \bibinfo{person}{Yichao Luo}, \bibinfo{person}{Yige Xu}, {and} \bibinfo{person}{Qi Zhang}.} \bibinfo{year}{2021}\natexlab{}.
\newblock \showarticletitle{One2Set: Generating Diverse Keyphrases as a Set}. In \bibinfo{booktitle}{\emph{{ACL/IJCNLP}}}. \bibinfo{pages}{4598--4608}.
\newblock


\bibitem[Yuan et~al\mbox{.}(2023)]%
        {DBLP:conf/acl/YuanH0023}
\bibfield{author}{\bibinfo{person}{Changsen Yuan}, \bibinfo{person}{Heyan Huang}, \bibinfo{person}{Yixin Cao}, {and} \bibinfo{person}{Yonggang Wen}.} \bibinfo{year}{2023}\natexlab{}.
\newblock \showarticletitle{Discriminative Reasoning with Sparse Event Representation for Document-level Event-Event Relation Extraction}. In \bibinfo{booktitle}{\emph{ACL}}. \bibinfo{pages}{16222--16234}.
\newblock


\bibitem[Zeng et~al\mbox{.}(2020)]%
        {DBLP:conf/coling/ZengJGGC20}
\bibfield{author}{\bibinfo{person}{Yutao Zeng}, \bibinfo{person}{Xiaolong Jin}, \bibinfo{person}{Saiping Guan}, \bibinfo{person}{Jiafeng Guo}, {and} \bibinfo{person}{Xueqi Cheng}.} \bibinfo{year}{2020}\natexlab{}.
\newblock \showarticletitle{Event Coreference Resolution with their Paraphrases and Argument-aware Embeddings}. In \bibinfo{booktitle}{\emph{{COLING}}}. \bibinfo{pages}{3084--3094}.
\newblock


\bibitem[Zhou et~al\mbox{.}(2022)]%
        {DBLP:conf/coling/ZhouDTWD22}
\bibfield{author}{\bibinfo{person}{Jie Zhou}, \bibinfo{person}{Shenpo Dong}, \bibinfo{person}{Hongkui Tu}, \bibinfo{person}{Xiaodong Wang}, {and} \bibinfo{person}{Yong Dou}.} \bibinfo{year}{2022}\natexlab{}.
\newblock \showarticletitle{{RSGT:} Relational Structure Guided Temporal Relation Extraction}. In \bibinfo{booktitle}{\emph{COLING}}. \bibinfo{pages}{2001--2010}.
\newblock


\end{thebibliography}

\appendix

\section{Dependency Parsing Chain}
\label{sec.parse_cot_construction}
\begin{figure}[htbp]
\centering 
\includegraphics[width=4.5cm]{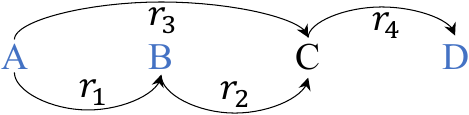} 
\caption{A, B, D represent event mentions, while C denotes other words. $r_1$, $r_2$, $r_3$, $r_4$ represent different dependency relations.} 
\label{Fig.parse_cot_construction} 
\end{figure}

We use the Stanford NLP toolkit's CoreNLP Dependency Parser to create a dependency parse tree from the context, generating various dependency edges. Their meanings are detailed in the toolkit's official documentation. In ERE tasks, we focus solely on the edges between event mentions and retain only the essential nodes and edges needed to connect them. If there is a tie in the number of nodes and edges, we keep them in the order they appear. As shown in Figure~\ref{Fig.parse_cot_construction}, both <$r_1$, $r_2$, $r_4$> and <$r_3$, $r_2$, $r_4$>. It is crucial to mention that since the dependency parser functions at the sentence level, we substitute "." with ";" to ensure the generation of the required dependency parsing chain.
 
\section{Experimental Settings}
\label{appendix.experimental_settings}
\noindent{\textbf{Dataset.}} Our experiments are conducted on four widely-used datasets (cf. Table~\ref{Tab.dataset}), including MAVEN-ERE~\citep{wang-etal-2022-maven} for unified event relation extraction, HiEve~\citep{DBLP:conf/lrec/GlavasSMK14} for sub-event relation extraction, MATRES~\citep{DBLP:conf/acl/RothWN18} for temporal relation extraction, and MECI~\citep{DBLP:conf/coling/LaiVNDN22} for causal relation extraction. 
For a fair comparison, we divided the data into the same training, validation, and test sets as in previous studies~\citep{wang-etal-2022-maven,hieu2022selecting,DBLP:conf/coling/ZhouDTWD22,DBLP:conf/coling/LaiVNDN22}. In particular, since the training and test sets are not separated, consistent with previous works, HiEve selects 80 documents for training (with a 0.4 probability for down-sampling negative examples) and 20 documents for testing. Since MAVEN-ERE does not have an open test set, we chose to use the validation set for testing.

\noindent{\textbf{Evaluation Metric.}} Based on previous research on event relation extraction~\citep{DBLP:conf/emnlp/ChoubeyH17a,DBLP:conf/emnlp/NguyenMDN22,DBLP:conf/eacl/WangZDGRC23,DBLP:conf/acl/YuanH0023,DBLP:conf/acl/CaselliV17,xu-etal-2022-extracting,DBLP:conf/naacl/NguyenMDN22}, we adopt the MUC~\citep{Vilain1995AMC}, B$^3$~\citep{bagga1998algorithms}, CEAF$_e$~\citep{Xiaoqiang1220579Luo} and BLANC~\citep{recasens_hovy_2011} metrics for evaluating event coreference relations. For the other three subtasks, we adopt the standard micro-averaged precision, recall, and F1 metrics. In particular, in the sub-event relation extraction task, PC and CP represent the F1 scores for parent-child and child-parent relations, respectively. 

\noindent{\textbf{Implementation Details.}}
MAQInstruct is conducted on a 4$\times$A100-80G setup. The input sequence length is 1536, and the output sequence length is 512. The weight for the bipartite matching loss, denoted as $\lambda$,  is set to 0.2. We use a learning rate of 5e-4, a batch size of 16, and a gradient accumulation of 2. The learning rate scheduler follows a cosine function, and the model is trained for 20 epochs. The results reported in the experiment are the averages of 5 different random seeds (0, 1, 2, 3, 4). We train the model using an Adam optimizer with weight decay, with the weight decay rate set to 1e-4. The warm-up proportion for the learning rate is 0.1, and the dropout rate is also 0.1. The temperature used to adjust the probabilities of the next token is set to 0.01, while the smallest set of the most probable tokens, with top\_p probabilities, adds up to 0.9. In the output, we use ":" as a delimiter to distinguish the dependency parsing chain from the multiple answers. BertERE employs RoBERTa as its backbone, utilizing a learning rate of 2e-5 for the transformer and 5e-4 for the classification MLP. It selects the longest text containing the event pair, constrained to 512 tokens. InstructERE approaches event relation extraction as a text generation task, employing the same backbone, pre-trained models, and training parameters as MAQInstruct.
\begin{table}[htbp]
\centering
\small
\caption{Dataset Statistics. "\#" denotes the amount. "Mentions" represents the potential events. "Links" means the event relations.}
\setlength\aboverulesep{0pt}\setlength\belowrulesep{0pt}
\begin{tabular}{lccc}
\toprule
Datasets & \#Docs   & \#Mentions & \#Links   \\ \hline
MAVEN-ERE    & 4,480 & 112,276 & 103,193 \\ \hline
HiEve    & 100    & 3, 185   & 3,648   \\ \hline
MATRES   & 275    & 11,861  & 13,573  \\ \hline
MECI     & 438    & 8,732    & 2,050    \\ \bottomrule
\end{tabular}
\label{Tab.dataset}
\end{table}

\section{Different Instructions Analysis}
\label{appendix.different_instruction}
The event relation extraction model based on LLMs is significantly influenced by the provided instructions. Experiments are conducted to assess various sets of instructions, revealing that for fixed tasks, shorter and more concise instructions are typically more effective. Additionally, multiple tests are performed, as outlined in Table~\ref{Tab.different_instruction}. Firstly, including all potential event mentions in the instructions results in a slight decline in the F1 score. Secondly, when the model is enabled to directly generate event relations based on event mentions, its performance significantly deteriorates due to the high volume of event mention pairs that produce relations labeled as NoRel. Furthermore, the model's performance reaches its lowest when multiple distinct relations are generated simultaneously.
\begin{table}[htbp]
\centering
\small
\caption{The F1 score of MAQInstruct on MECI varies among different instructions.}
\setlength\aboverulesep{0pt}\setlength\belowrulesep{0pt}
\begin{tabular}{m{7.0cm}|m{0.7cm}<\centering}
\toprule
Instruction                                              & MECI                  \\ \hline
List the \textbf{\textit{cause}} event of   <0x85> \textcolor{black}{\textbf{\textit{earthquake}}} ?            & 62.3 \\ \hline
Find the \textbf{\textit{cause}} event of <0x85> \textcolor{black}{\textbf{\textit{earthquake}}} from the event mentions <0x71> \textbf{\textit{scorched}}, …? & 61.7 \\ \hline
What's the event relation between <0x85> \textcolor{black}{\textbf{\textit{earthquake}}} and <0x71> \textbf{\textit{scorched}}, <0x72> \textbf{\textit{deny}}, …?      & 60.4 \\ \hline
List the \textbf{\textit{cause}} and \textbf{\textit{effect}} event of  <0x85> \textcolor{black}{\textbf{\textit{earthquake}}} ? & 56.6 \\ \bottomrule
\end{tabular}
\label{Tab.different_instruction}
\end{table}

\section{Different Markers Analysis}
\label{appendix.different_marker}
In this study, we employ various markers to elicit event mentions, building upon prior research~\cite{lu-etal-2022-unified,ye-etal-2022-packed}. We categorize the experiments into three distinct groups, as illustrated in Table~\ref{Tab.different_marker}. The first group utilizes special tokens from Llama2, such as <0x**>, which yield optimal results. Notably, the addition of a special end character following event mentions does not enhance performance due to insufficient semantic information and the introduction of multiple tokens that compromise coherence. The second group substitutes <0x**> with <No**>, resulting in a significant reduction in effectiveness, as the presence of excessive tokens again leads to incoherence. In the third group, the application of a uniform marker for all event mentions produces markedly poorer outcomes.
\begin{table}[htbp]
\centering
\small
\caption{The F1 score of MAQInstruct on MECI varies among different markers.}
\setlength\aboverulesep{0pt}\setlength\belowrulesep{0pt}
\begin{tabular}{m{1.25cm}|m{4.4cm}|m{0.65cm}<\centering}
\toprule
Marker             & Tokenizer                                       & MECI \\ \hline
<0x64>             & [103]                                         & 62.3 \\ \hline
<0x64> </>         & \leftline{[103]}[1533, 29958]              & 62.1 \\ \hline
<No64>             & [529, 3782, 29953, 29946, 29958]              & 59.5 \\ \hline
<No64> </>         & \leftline{[529, 3782, 29953, 29946, 29958]}[1533, 29958] & 59.3 \\ \hline
<Strong>           & [529, 1110, 29958]                            & 60.2 \\ \hline
<Strong> </> & \leftline{[529, 1110, 29958]} [1533, 29958]   & 59.7 \\ \bottomrule
\end{tabular}
\label{Tab.different_marker}
\end{table}

\section{Case Study}
\begin{figure}[h]
    \centering 
    \includegraphics[width=7.7cm]{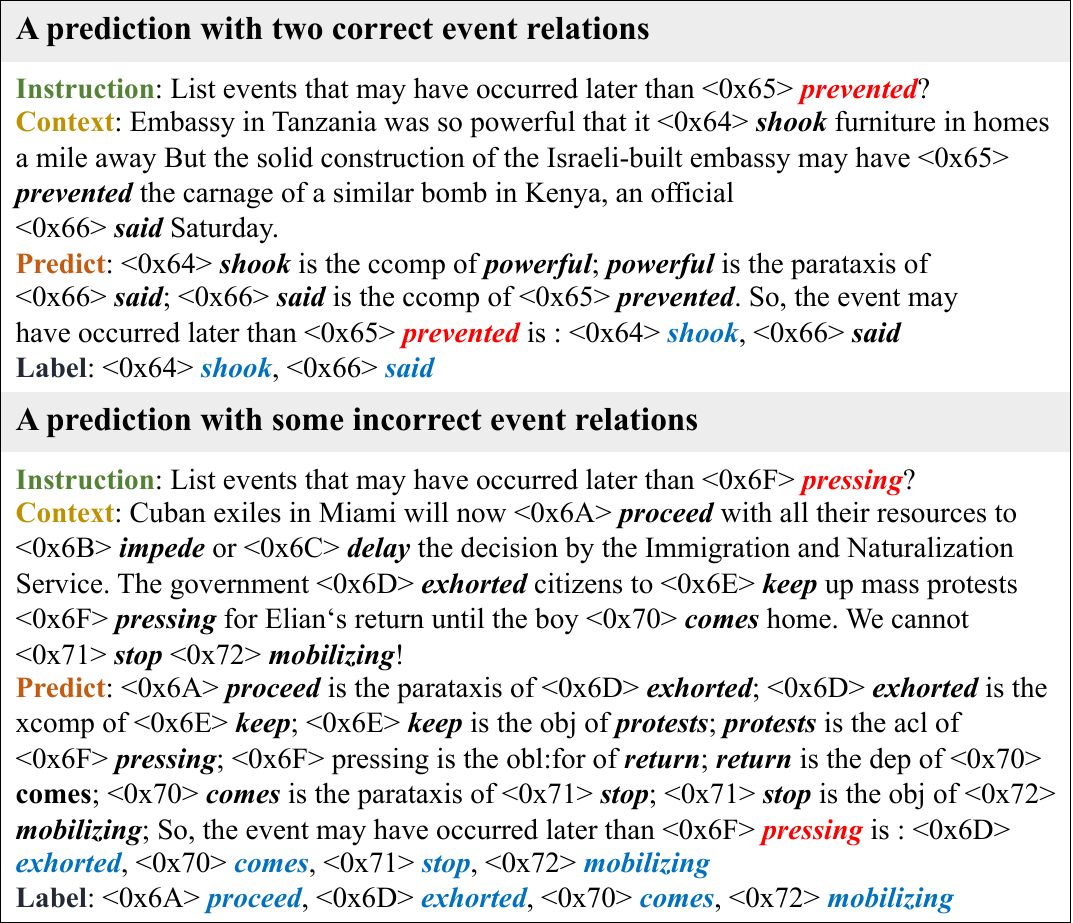} 
    \caption{Two examples demonstrating the use of MAQInstruct in extracting temporal relations.} 
    \label{Fig.case_study}
\end{figure}

This study presents a qualitative analysis of the extraction of multiple answers, utilizing two examples of event temporal relation extraction, as illustrated in Figure~\ref{Fig.case_study}. The first example effectively demonstrates the extraction process, supported by a valuable dependency parsing chain. 
Conversely, the erroneous example reveals two issues: the failure to recall event <0x6A> and the incorrect recall of event <0x71>. These inaccuracies arise from the complexity of the dependency parsing chain, which obscures pertinent structural information while introducing extraneous details, resulting in errors.

\section{Related Work}
Previous methods for event relation extraction~\cite{hieu2022selecting,hwang-etal-2022-event,huang2023classification,DBLP:conf/acl/BarhomSEBRD19,DBLP:conf/acl/HuLJ0GGC23,wang-etal-2022-maven,DBLP:conf/eacl/TanPH23} primarily utilize multi-class classification, MASK prediction, or prototype matching, focusing on addressing specific sub-tasks such as coreference, temporal, causal, or sub-event relations. In the classification-based approach~\cite{huang2023classification,DBLP:conf/naacl/LuN21,DBLP:conf/acl/TranPN20,DBLP:conf/coling/ZengJGGC20,DBLP:conf/emnlp/WangCZR20,DBLP:conf/acl/BarhomSEBRD19}, event mentions are paired together, and additional features are incorporated, such as prototypes, logical rules, graph convolutional networks, or prompts. MASK prediction-based methods~\cite{DBLP:journals/corr/abs-2307-09813,DBLP:conf/coling/ShenZWQ22,DBLP:conf/coling/CuiSCLLS22} train a masked language model to predict the relation. The prototype matching method~\cite{hu2023protoem} manually selects instances to serve as prototypes for each relation, and new instances are then matched against these prototypes. \citet{DBLP:conf/emnlp/SegalESGB20} and \citet{hu-etal-2019-multi} each proposed reading comprehension models based on multi-choice and multi-span, respectively, allowing the model to select the correct answer from candidate options or to generate multiple answers simultaneously. Simultaneously, many entity relation extraction methods based on LLMs~\cite{wang2023instructuie,xiao2024yayiuie} directly prompt large language models to generate relations between pairs of entities. However, these methods have several drawbacks. Therefore, we have designed a series of improvement measures to address these identified deficiencies.

\end{document}